\documentclass{llncs}

\usepackage{floatrow}
\usepackage{url}
\usepackage{hyperref}
\usepackage{enumerate}
\usepackage{graphicx}
\usepackage{subcaption}
\usepackage[strings]{underscore}
\usepackage{color}
\usepackage{array}
\usepackage[font=small,labelfont=bf,tableposition=top]{caption}

\newcolumntype{M}[1]{>{\centering\arraybackslash}m{#1}}

\begin{document}

\title{License Plate Detection and Recognition Using Deeply Learned Convolutional Neural Networks}

\author{Syed Zain Masood \hspace{0.25 in} Guang Shu \hspace{0.25 in} Afshin Dehghan \hspace{0.25 in} Enrique G. Ortiz \\
		{\tt\small \{zainmasood, guangshu, afshindehghan, egortiz\}@sighthound.com}}

\institute{Computer Vision Lab, Sighthound Inc., Winter Park, FL}

\maketitle              

\begin{abstract}
This work details Sighthound\'s fully automated license plate detection and recognition system. The core technology of the system is built using a sequence of deep Convolutional Neural Networks (CNNs) interlaced with accurate and efficient algorithms. The CNNs are trained and fine-tuned so that they are robust under different conditions (e.g. variations in pose, lighting, occlusion, etc.) and can work across a variety of license plate templates (e.g. sizes, backgrounds, fonts, etc). For quantitative analysis, we show that our system outperforms the leading license plate detection and recognition technology i.e. ALPR on several benchmarks. Our system is available to developers through the Sighthound Cloud API at \url{https://www.sighthound.com/products/cloud}
\end{abstract}

\section{Introduction}
In the past, there has been considerable focus on license plate detection \cite{Hsieh02,Arth07,Yuan17} and recognition techniques \cite{Du13,Li16,Chang04,Anagnostopoulos08,Gou16,Cheang17,Hsu13}. From traffic and toll violations to accident monitoring, the ability to automatically detect and recognize plates is one of the key tools used by law enforcement agencies around the world. 

Contrary to popular belief, license plate detection and recognition is still a challenging problem due to the variability in conditions and license plate types. Most of the existing solutions are restrictive in nature i.e. work for stationary cameras, with a specific viewing angle, at a specific resolution, for a specific type of license plate template. What this means is that a license plate system designed for a static camera in Europe for example will perform poorly for a moving camera in the US. Additionally, because of processing speed bottlenecks, most techniques compare and pick the best possible feature and rely heavily on heuristic methods. This is less than ideal in a world where a combination of different types of features has shown to perform better than selecting a single feature for a given task.

With the advent \cite{LeCun90,LeCun04,Jarrett09,Krizhevsky12,Lee09} and enhancements \cite{Vmmcr17,Dager17} of deep CNNs and cheaper, faster processing hardware, we need to take a fresh look at the problem. Advances in the design of CNNs have resulted in significant increases in performance accuracies for many tasks. In this work, we harness the power of CNNs in an end-to-end system capable of detecting and recognizing license plates with low error rates. We present a system that is robust to variations in conditions (camera movement, camera angle, lighting, occlusion, etc) and license plate templates (size, designs, formats, etc). In order to support our claims, we tested our system on several benchmarks and achieved results better than the previous state-of-the-art. The contributions of this work are summarized below. 

\begin{itemize}
\item We present an end-to-end license plate detection and recognition pipeline, along with novel deep CNNs, that not only are computationally inexpensive, but also outperform competitive methods on several benchmarks.
\item We conducted a number of experiments on existing benchmarks and obtained leading results on all of them.
\end{itemize}

\begin{figure}
\begin{center}
   \includegraphics[width=\linewidth]{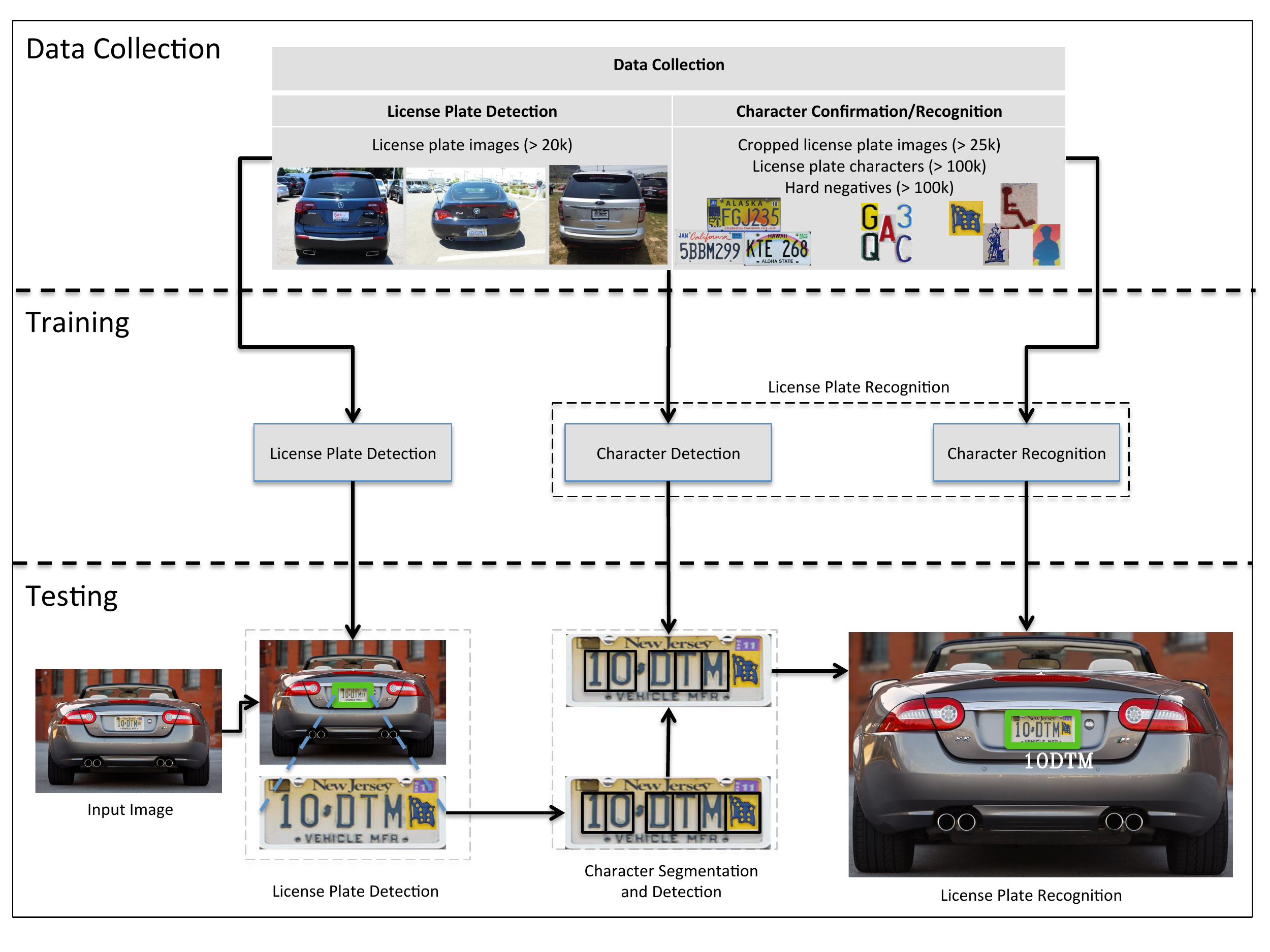}
   \caption{This figure shows the complete pipeline of our license plate recognition system. As observed, the process is divided into three main categories: 1) Data collection 2) Deep CNN training 3) Testing. The collected data is cleaned, organized and properly annotated (in a semi-automated manner) before being used to train deep models for license plate detection, character detection and character recognition. When testing an image: 1) The license plate detector is used to find and crop the license plate followed by; 2) character segmentation to isolate each character before; 3) using the character detection module to filter out non-character elements, and; 4) running each remaining character through the recognition system to get the final license plate number.}
   \label{figPipeline}
\end{center}
\end{figure}

\section{System Overview}
Figure \ref{figPipeline} presents an overview of our system. As can be seen, there are three main components of our approach, namely data collection, training and testing. Below, we talk about each of these components in detail.

\subsection{Data Collection and Preparation}
To train accurate and efficient deep CNNs, it is vital to have significant amounts of clean and correctly annotated data. For this purpose, we spend time and energy to make sure the collected data to be used for training is well organized. As shown in Figure \ref{figPipeline}, we need to collect data for two tasks: license plate detection and character detection/recognition.

\subsubsection{Data: License Plate Detection}
For the task of license plate detection, the need is to construct a database of real world images with license plates, along with context (vehicle, scene, etc), clearly visible. To build a system that is robust to variations in environment and license plate specifications, it is important to collect a set of images that exhibits such cases. Hence, for this task we collected more than 20k images from different sources capturing all needed variations. A few examples of the type of images required are shown in the 1st column of the data collection table in Figure \ref{figPipeline}. Once we had a database of these images, we devised a semi-automated technique to clean, annotate and prepare the images for training.

\subsubsection{Data: License Plate Recognition}
In order to train a network to recognize license plate characters, we set out to collect image data of tightly cropped license plates (no context). Like the real world license plate data collection process above, we made sure to collect data with different backgrounds, fonts, and under different lighting conditions. With this in mind, we were able to collect 25k images of cropped license plates. We then used an automated process to segment, extract and categorize characters, along with hard negative background graphic samples, before manually cleaning the data. A few samples of each (cropped license plates, characters and hard negatives) are shown in the 2nd column of the data collection table in Figure \ref{figPipeline}.

\begin{figure}
\centering
\begin{subfigure}{.5\textwidth}
  \centering
  \includegraphics[width=0.95\linewidth]{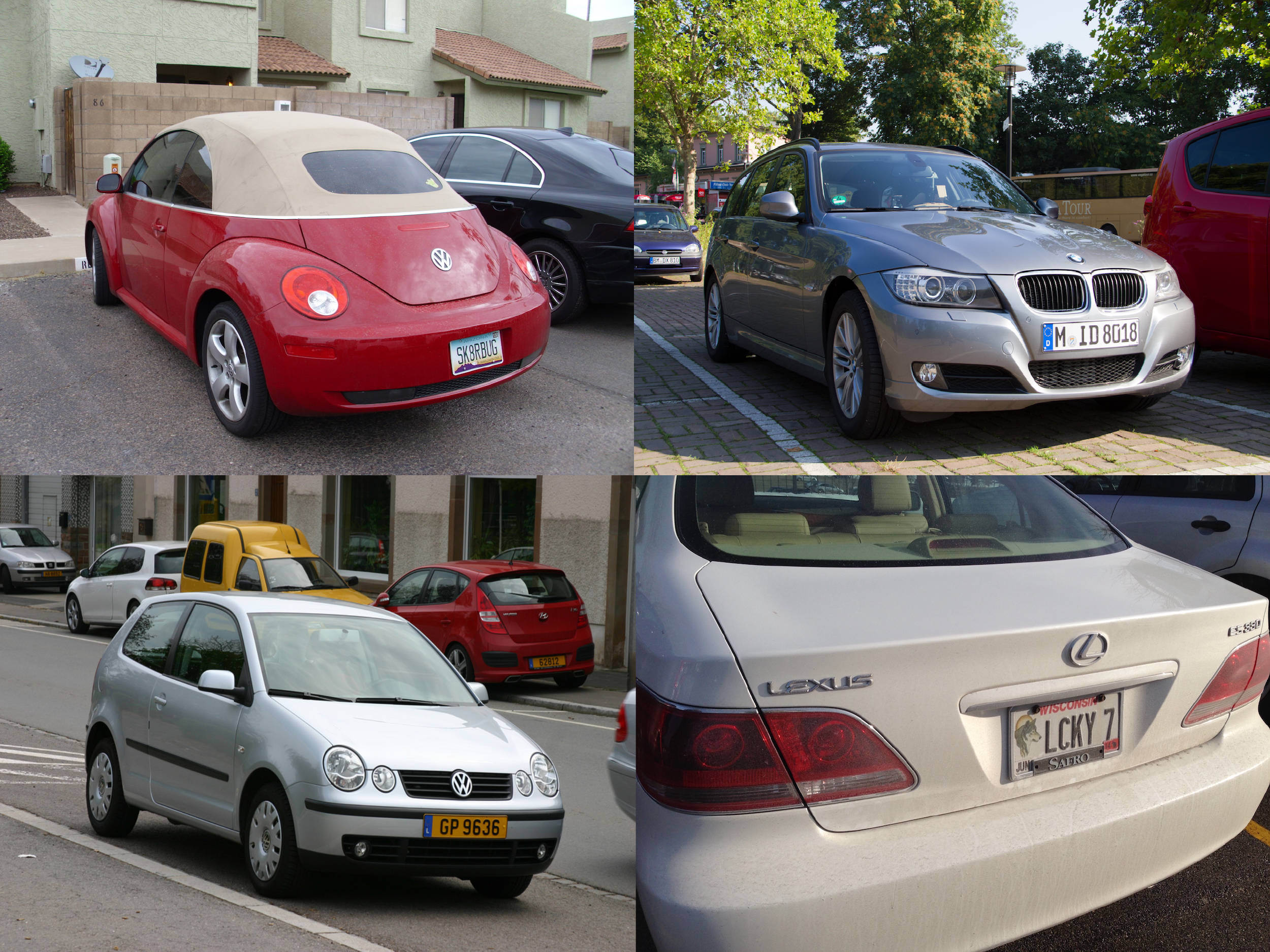}
  \caption{}
  \label{fig:sub1}
\end{subfigure}%
\begin{subfigure}{.5\textwidth}
  \centering
  \includegraphics[width=0.95\linewidth]{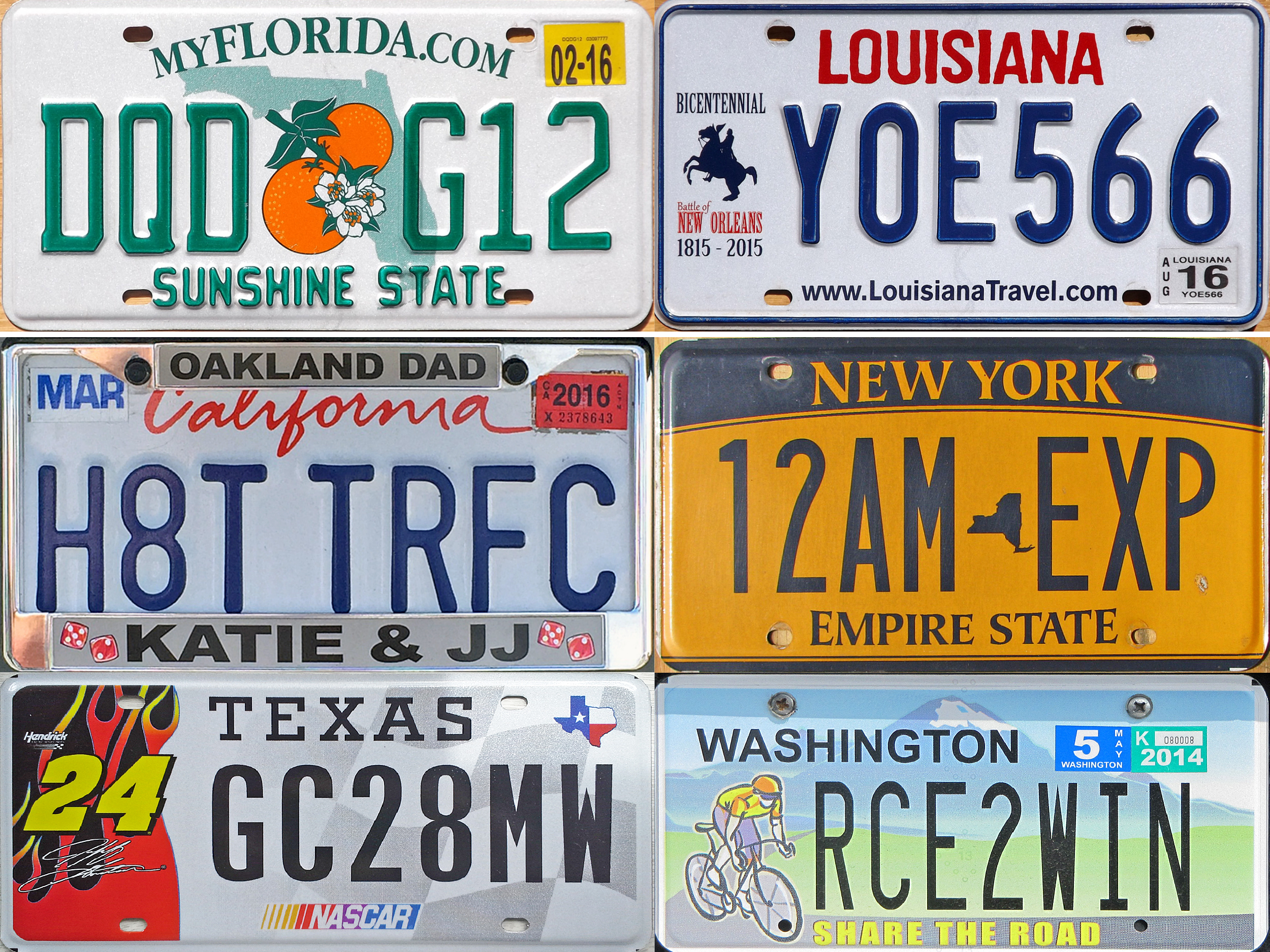}
  \caption{}
  \label{fig:sub2}
\end{subfigure}
\caption{Figure shows (a) data samples of real world license plate images used to train the license plate detector and (b) data samples of tightly cropped license plates from which characters are extracted to train character detection and recognition models.}
\label{fig:test}
\end{figure}

\subsection{Training}
Once the data is prepared and ready to be used for training, we train three separate deep CNNs: license plate detection, character detection and character recognition.

\subsubsection{Training: License Plate Detection} Similar to \cite{Vmmcr17}, train a deep CNN for the task of license plate detection. This model is responsible for localizing and classifying a license plate within a given image. For training, we use a database of real world license plate images along with license plate annotations to train this network. Since the training data we collected covers a wide range of variations, the network is designed to adapt to varying situations and be robust to regional differences in license plates.  

\subsubsection{Training: License Plate Recognition} Train a series of CNNs for the following tasks:
\begin{itemize}
\item Confirm if a character exists. It is a binary deep network classifier trained with license plate character segments as positives and plate backgrounds and symbols as negatives. The variety of character fonts and healthy samples of hard negative samples help in training a robust system. Hence, this model helps filter out non-character segments for the character recognition task to follow. 
\item If characters exist, recognize the character. On the same lines as \cite{Dager17}, train a deep CNN for the task of character recognition. The network is trained with the complete character dataset comprising $35$ different classes i.e. $0$-$9$ and A-Z excluding O (Due to the similarity of \textit{O} and \textit{0}, we consider them the same). Lower case characters are ignored since license plates do not contain them. Font, lighting and crop variations of character segments generate a network capable of working across different situations.
\end{itemize}


\subsection{Testing}
The workings of our automatic license plate detection and recognition pipeline is illustrated in the testing section (3rd row) of Figure \ref{figPipeline}. Below we explain each of the steps involved in detail.

\subsubsection{Testing: License Plate Detection}
Given an input image, the first task is to determine if it contains a license plate. To achieve this task, we run the image through our deep license plate detection network. As explained earlier, this network is robust to different conditions (e.g. lighting changes, viewing angles, occlusion, etc) and license plate variations (e.g. shape, format, background, etc), making it highly accurate across a range of scenarios.


\subsubsection{Testing: License Plate Recognition} Once the license plate is detected, we need a mechanism by which we can read the plate. The best approach is to segment/isolate each of the license plate characters, perform character detection and recognition for each individual character before combining the results. 

To achieve this goal, we devised a number of segmentation algorithms and techniques that are able to work seamlessly across known variations. In order to ensure all license plate characters are captured, we designed a multi-step segmentation process. The level of complexity for character segmentations increases with each step; the initial step captures simple character segmentations while the later steps help isolate difficult/obscure characters. It is also helpful to note that we generally keep the segmentation checks lenient since we used the next step to help filter characters from non-characters. 

Since we use lenient techniques for segmenting characters, there are bound to be non-character segments in the final segment list. We help filter these out by using a character detection deep network that is trained using true license plate characters against non-character symbols (e.g. wheelchair, flags, etc). This binary classification adds an extra layer of correctness by excluding non-character segments, that might look similar to a character, from the final recognition task. As a final step, each character segment is classified into one of $35$ classes before constructing the license plate result.

\section{Experiments}
This section details the experimental results on benchmark datasets along with comparisons with current state-of-the-art technology.

\subsection{Datasets}
To benchmark our pipeline, we tested against publicly available license plate datasets. To show that our system is agnostic to license plates from different regions, we selected two datasets for USA license plates \cite{CaltechCars,OpenALPRUS} containing $348$ images and a collection of $608$ images of European license plates \cite{OpenALPREU,CroatiaPlates}

\begin{figure}
\begin{center}
   \includegraphics[width=\linewidth]{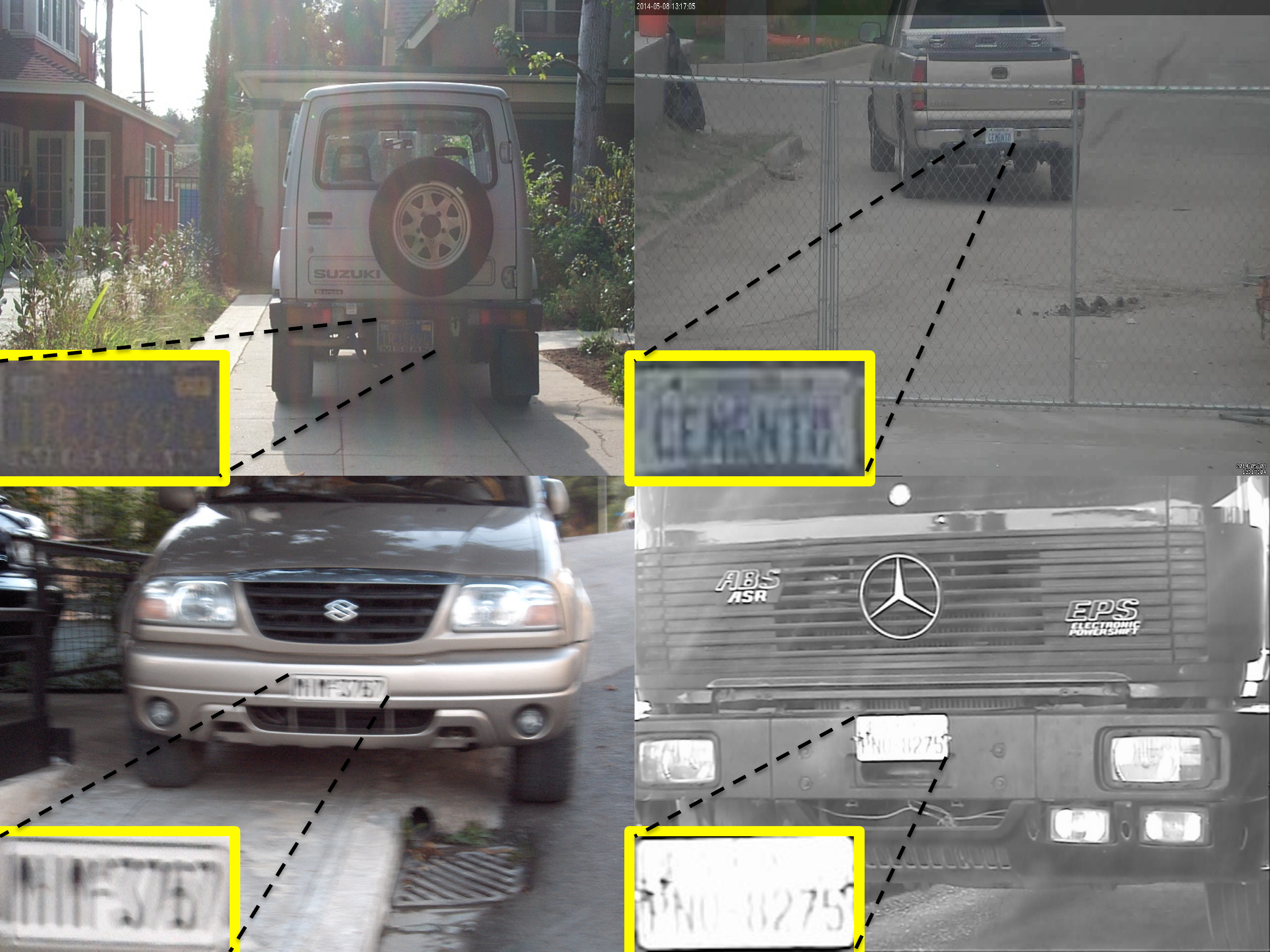}
   \caption{Figure shows some of the hard examples (lighting, blurriness, size, angle, etc) removed before running benchmarks. The cropped license plate used for recognition processing is shown in the bottom-left corner for each sample.}
   \label{fig:hardexamples}
\end{center}
\end{figure}

In order to be able to read the license plate, processing is done at a certain resolution. Although most of the images in these datasets are reasonable, there are a few exceptions where illumination, size and blurriness make it extremely hard to find and/or read the plate. Some samples of these hard examples are show in Figure \ref{fig:hardexamples}. To be fair to the recognition systems, we removed these examples before performing the benchmark testing \footnote{List of all removed samples can be found in the supplementary material}. It should be noted that removal of these unreasonably hard samples does not imply that the remaining images are all clean and simple. The data still contains a variety of images, both simple and hard examples with variations in lighting, pose, size, etc. The updated dataset information is as follows:
\begin{itemize}
\item $328$ images of US license plates after removing $20$ samples
\item $550$ images of European license plates after removing $58$ samples
\end{itemize}

\subsection{Comparison}
For comparison, we benchmark our results against OpenALPR open source \cite{OpenALPROpenSource} and OpenALPR Cloud API \cite{OpenALPRCloud}. 
It is worth noting that unlike OpenALPR, Sighthound's method does not require user input for the type of plate (e.g. USA, Europe). This means that a single unified model is good for plates from these region.

Since the goal of this paper is to benchmark an end-to-end system, failure to detect a license plate results in a penalty at the recognition stage as well. Hence, any plates we fail to detect will imply a failure to recognize the plate.

\subsection{Results}
Here we present both quantitative and qualitative results of our pipeline against the solutions provided by OpenALPR \footnote{Tests conducted Feb 2017} across different datasets for license plate detection and recognition tasks.

\subsubsection{Quantitative Result: License Plate Detection}


We use precision and recall to compare different methods. Since all techniques achieve 100\% precision, the meaningful comparison lies in the recall numbers. 

Table \ref{tab:detection} shows the recall results of our license plate detector compared to the OpenALPR competitors on different datasets. We can clearly see that Sighthound's license plate detector performs much better than the solutions proposed by OpenALPR. Hence, our system is almost always is able find the license plates in a given scene.
\begin{table}[ht!]
\caption{License plate detection recall numbers across different datasets.}
\centering
\begin{tabular}{ |M{2.5cm}|M{2.5cm}|M{2.5cm}|M{2.5cm}| } 
\hline
Dataset & OpenALPR (Open Source) & OpenALPR (Cloud API) & Sighthound \\ 
\hline\hline
USA & 86.89\% & 89.33\% & \color{red}\textbf{99.09}\% \\
Europe & 91.09\% & 90.36\% & \color{red}\textbf{99.64}\% \\
\hline
\end{tabular}
\label{tab:detection}
\end{table}

\subsubsection{Quantitative Result: License Plate Recognition}
We evaluate both methods using a best sequence match technique; the most common sequence between the predicted result and ground truth is used to compute the accuracy of the license plate. Table \ref{tab:recognition} present results of our recognition method compared to OpenALPR. As expected, the license plate recognition approach proposed by Sighthound improves on both OpenALPR implementations. It is commendable that our system, with all possible variations in the data, is able to recognize the license plate string correctly with an accuracy of greater than $90\%$.


\begin{table}
\caption{License plate recognition accuracies across different datasets.}
\centering
\begin{tabular}{ |M{2.5cm}|M{2.5cm}|M{2.5cm}|M{2.5cm}| } 
\hline
Dataset & OpenALPR (Open Source) & OpenALPR (Cloud API) & Sighthound \\ 
\hline\hline
USA & 78.36	\% & 84.64\% & \color{red}\textbf{93.44}\% \\
Europe & 84.80\% & 86.75\% & \color{red}\textbf{94.55}\% \\
\hline
\end{tabular}
\label{tab:recognition}
\end{table}

\subsubsection{Qualitative Results}
\begin{figure}
\begin{center}
   \includegraphics[width=0.95\linewidth]{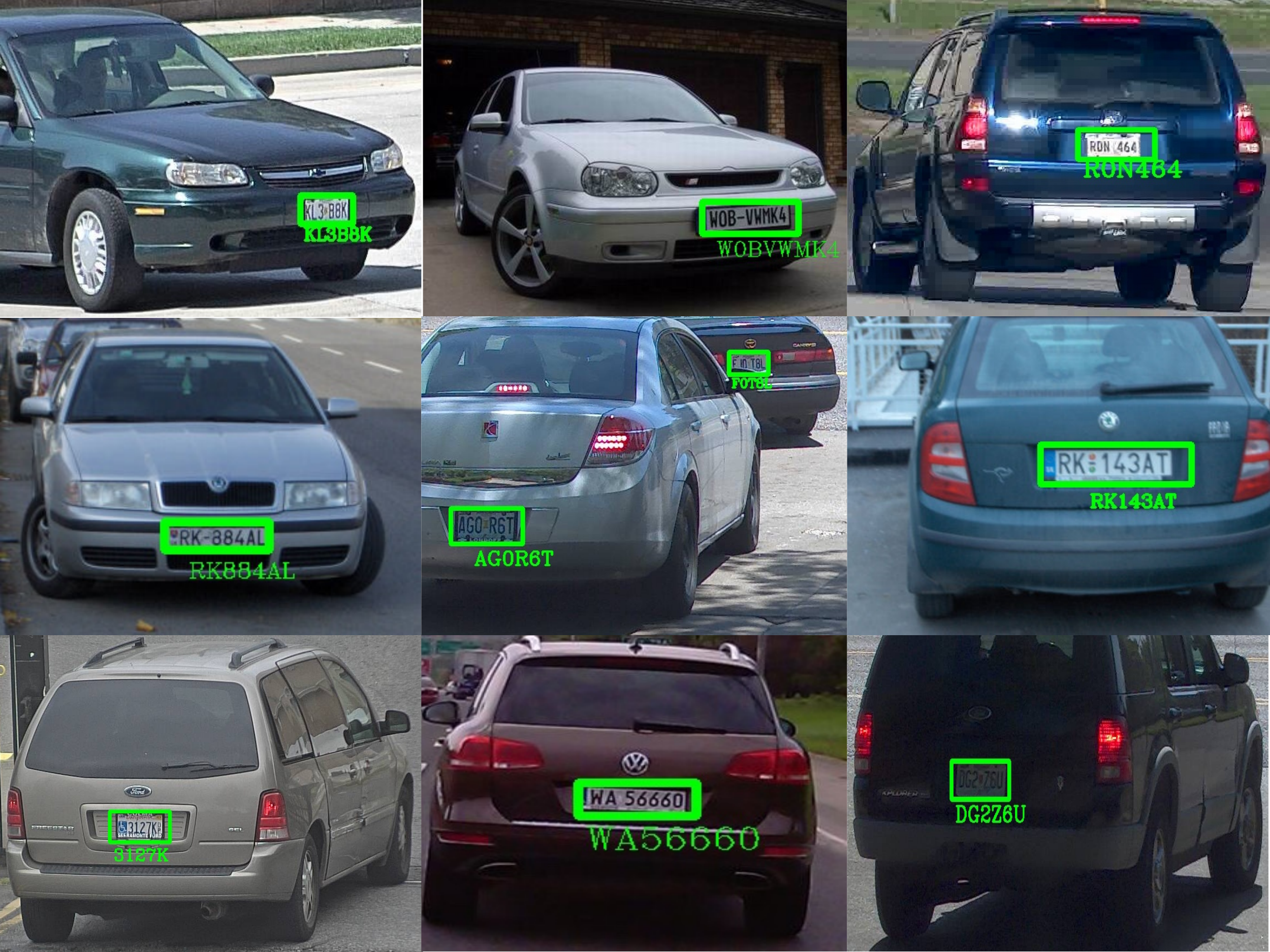}
   \caption{Figure shows some results of our end-to-end license plate detection and recognition system on plates from USA, Europe and Brazil. One can observe that our unified pipeline is indifferent to plates from different regions and is robust to variations in illumination, pose, size, blurriness, etc.}
   \label{fig:results}
\end{center}
\end{figure}

Figure \ref{fig:results} Here we present some qualitative results using our license plate detection and recognition system. One can observe that the system works well across a wide range of variations (e.g. illumination, size, blurriness, angle, etc) and license plate templates (e.g. USA, Europe, Brazil, etc).

\section{Conclusion}
In this paper, we present an end to end system for license plate detection and recognition. We introduce a novel pipeline architecture, comprising a sequence of deep CNNs, to address the detection and recognition tasks and show that our pipeline outperforms state-of-the-art commercial solutions on benchmark datasets.

%
%
\bibliographystyle{splncs}
\bibliography{egbib}

\end{document}